\documentclass{llncs}

\usepackage{graphicx}
\usepackage{graphicx, caption}
\usepackage{subfig}
\usepackage{cleveref}

\begin{document}

\title{ResQbot: A Mobile Rescue Robot \\ with Immersive Teleperception \\ for Casualty Extraction}

\titlerunning{Mobile Rescue Robot}  % abbreviated title (for running head)
\author{Roni Permana Saputra\inst{1,}\inst{2} \and Petar Kormushev\inst{1}}
\authorrunning{Roni Permana Saputra et al.} % abbreviated author list (for running head)
\tocauthor{Roni Permana Saputra and Petar Kormushev}
\institute{Robot Intelligence Lab, Dyson School of Design Engineering, \\Imperial College London, United Kingdom\\
\email{\{r.saputra,p.kormushev\}@imperial.ac.uk}\\ 
\texttt{https://www.imperial.ac.uk/robot-intelligence/}
\and
Research Centre for Electrical Power and Mechatronics,\\
Indonesian Institute of Sciences - LIPI, Indonesia\\ 
%\texttt{http://www.telimek.lipi.go.id/} 
}

\maketitle

\begin{abstract}
In this work, we propose a novel mobile rescue robot equipped with an immersive stereoscopic teleperception and a teleoperation control. This robot is designed with the capability to perform safely a casualty-extraction procedure. We have built a proof-of-concept mobile rescue robot called ResQbot for the experimental platform. An approach called "loco-manipulation" is used to perform the casualty-extraction procedure using the platform. The performance of this robot is evaluated in terms of task accomplishment and safety by conducting a mock rescue experiment. We use a custom-made human-sized dummy that has been sensorised to be used as the casualty. In terms of safety, we observe several parameters during the experiment including impact force, acceleration, speed and displacement of the dummy's head. We also compare the performance of the proposed immersive stereoscopic teleperception to conventional monocular teleperception. The results of the experiments show that the observed safety parameters are below key safety thresholds which could possibly lead to head or neck injuries. Moreover, the teleperception comparison results demonstrate an improvement in task-accomplishment performance when the operator is using the immersive teleperception.  

\end{abstract}

\section{Introduction}

Catastrophic events, disasters, or local incidents generate hazardous and unstable environments in which there is an urgent need for timely and reliable intervention, mainly to save lives. A multi-storey building fire disaster---such as the recent Grenfell Tower inferno in London, United Kingdom~\cite{6:grenfell-tower}---is an example of such a scenario. Responding to such situations is a race against time---immediate action is required to reach all potential survivors in time. Such responses, however, are limited to the availability of trained first responders as well as prone to potential risks to their lives. The high risk to the lives of rescue workers means that, in reality, fast response on-site human intervention may not always be a possibility.

Using robots in search-and-rescue (SAR) missions offers a great alternative by potentially minimising the danger for the first responders. Moreover, it is more flexible---the number of these robots can be expanded to perform faster responses. Thus, various robotic designs have been proposed to suit several specific SAR applications~\cite{11:yoo2011military,2:Murphy:2008SAR}. These robots are designed to assist with one or sometimes multiple tasks as part of a SAR operation---including reconnaissance, exploration, search, monitoring, and excavation.

\begin{figure}[!t]
    \centering
        \subfloat[ResQbot robot platform] {\includegraphics[width=0.5\columnwidth]{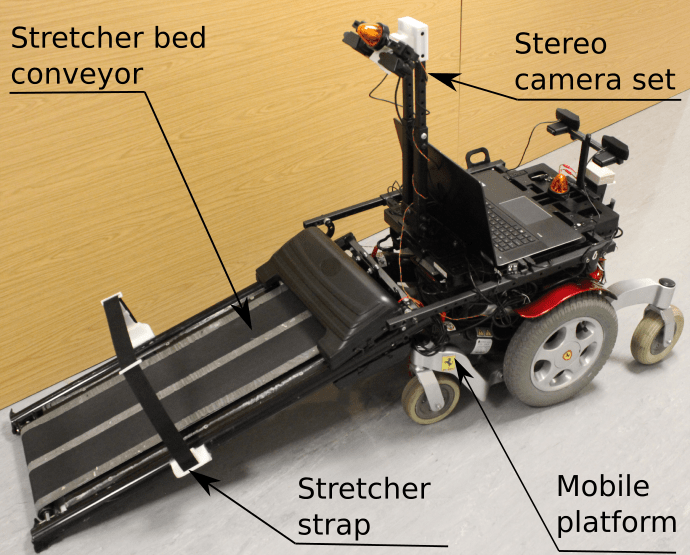}}
        \label{fig:resqbot-platform}
         \hspace{0.1cm}
        \subfloat[Teleoperating ResQbot]{\includegraphics[width=0.445\columnwidth]{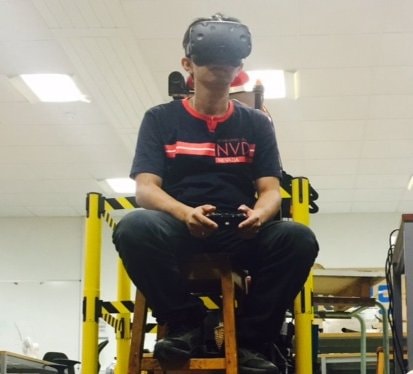}}
        \label{fig:resqbot-tele}
    \caption{Proposed mobile rescue robot with immersive stereoscopic teleperception: (a) The ResQbot platform comprising a motorised stretcher bed conveyor attached to a differential-drive mobile base, a stretcher strap module and a stereo camera rig, (b) Operator provided with teleperception via HTC-Vive headset.}
    \label{fig:prototype}
\end{figure}

Our aim in this study is to develop a mobile rescue robot system (see Fig.~\ref{fig:prototype}a) that is capable of performing a casualty-extraction procedure. This procedure includes loading and transporting a human victim---a.k.a. casualty---smoothly, which is essential for ensuring the victim's safety, via teleoperation mode. We also aim to develop more immersive teleperception for the robot's operators to improve their performance and produce higher operation accuracy and safer operations.

\section{Related Work}

Wide-ranging robotics research studies have been undertaken in the area of search, exploration, and monitoring, specifically with applications in SAR scenarios~\cite{3:Shen:2012uavindoor,4:Waharte:2010supportingUAVs,5:Goodrich:2008supporting}. Despite the use of the term 'rescue' in SAR, little attention has been given to the development of a rescue robot that is capable of performing a physical rescue mission, including loading and transporting a victim to a safe zone---a.k.a. casualty extraction.

Several research studies have been conducted to enable the use of robots in the rescue phase of SAR missions. The majority of these studies focused on developing mobile robots---mainly tracked mobile robots---with mounted articulated manipulators~\cite{7:Telemax:online,8:Schwarz:2017nimbro} or other novel arm mechanisms, such as elephant trunk-like arms~\cite{9:Wolf:2003elephant} or snake-like arms~\cite{2:Murphy:2008SAR}. Such designs enable the mobile robots to move debris by using their manipulators and also perform other physical interventions during the rescue mission, such as a casualty-extraction procedure.

Battlefield Extraction Assist Robot (BEAR) is one of the most sophisticated robot platforms designed and developed specifically for casualty-extraction procedures~\cite{11:yoo2011military,10:theobald2010mobile}. This robot was developed by Vecna Technologies and intended for the U.S. Army~\cite{12:robotics2010bear}. It has a humanoid form with two independent, tracked locomotion systems. It is also equipped with a heavy-duty, dual-arm system that is capable of lifting and carrying up to 227-kg loads~\cite{13:newatlas2010bear}. This robot is capable of performing casualty-extraction procedures by lifting up and carrying the casualty using its two arms. However, such a procedure could cause additional damage to the already injured victim---such as a spinal or neck injury from the lack of body support during the procedure.

Several investigations have been conducted on the development of rescue robots capable of performing safer and more robust casualty-extraction routines. These robots were designed to be more compact by using stretcher-type constructions or litters~\cite{14:6106797,15:5981556,16:6343761,17:sahashi2011study}. The robots presented in these studies are intended for performing safe casualty extractions with simpler mechanisms compared to arm mechanisms. The use of the stretcher-type design on this robot also ensures the victim's safety during transportation.

The mobile rescue robot demonstrated by the Tokyo Fire Department is one of the robots that uses this design concept~\cite{18:ota2011robocue,19:tele-rescue}. This robot is equipped with a belt-conveyor mechanism and also a pair of articulated manipulators. It uses its manipulators to lift up the casualty and place it onto the conveyor during the casualty-extraction procedure~\cite{20:popular-rescue}. Then the belt conveyor pulls up the casualty into the container inside the robot for safe transportation. Compared to BEAR, this robot offers a safer casualty-transportation process. However, lifting up the casualty during the loading process is still a procedure that is highly likely to cause additional damage to the victim.

Iwano et al. in~\cite{14:6106797,15:5981556,16:6343761} proposed a mobile rescue robot platform capable of performing casualty extraction without a "lifting" process. This robot loads the casualty merely using a belt-conveyor mechanism. This belt conveyor pulls the casualty from the ground onto the mobile platform, while the mobile platform synchronises the movement toward the body [15]. Since there is no "lifting" process during the casualty-extraction procedure, this method is expected to be safer than the methods applied in BEAR and the Tokyo Fire Department's robot. However, to the best of our knowledge, no safety evaluation of performing this extraction method has been published.

In terms of controlling method, these robots are still manually operated or teleoperated by human operators. The BEAR robot and the rescue robot demonstrated by the Tokyo Fire Department are teleoperated by human operators using a conventional teleoperation control setup~\cite{19:tele-rescue}. On the other hand, based on the reports presented in~\cite{14:6106797,15:5981556,16:6343761}, the rescue robot developed by Iwano et al. still requires human operators to be present on the scene to perform casualty-extraction procedures.

\section{Research Contributions}
In this work, we present a mobile rescue robot we developed that is capable of performing a casualty-extraction procedure using the method presented in~\cite{15:5981556}. Moreover, we equip this robot with teleoperation control and an immersive stereoscopic teleperception. As a proof of concept, we have designed and built a novel mobile rescue robot platform called ResQbot~\cite{21:resqbot} shown in Fig.~1. This robot is equipped with an onboard stereoscopic camera rig to provide immersive teleperception that is transmitted via a virtual-reality (VR) headset to the operator.

The contributions to this work are:
\begin{enumerate}
\item Proof-of-concept ResQbot, including teleoperation mode with immersive teleperception via HTC Vive\footnote{Virtual Reality Head Mounted Display by VIVE: https://www.vive.com/} headset;
\item Preliminary evaluation of loco-manipulation-based casualty extraction using the ResQbot platform, in terms of task accomplishment and safety;
\item Evaluation of the proposed immersive teleperception compared with the other teleperception methods, including the conventional teleperception setup and direct observation as a baseline.
\end{enumerate}

\section{Casualty Extraction via Loco-Manipulation Approach}

To perform a casualty-extraction procedure using the ResQbot platform, we propose using a loco-manipulation approach. By using this approach, the robot can implicitly achieve a manipulation objective---which is loading of a victim onto the robot---through a series of locomotive manoeuvres. We utilise the conveyor module mounted on the mobile robot base to create a simple mechanism for the loading of a victim onto the conveyor surface by solely following a locomotive routine.

\begin{figure}[t!]
\centering
\includegraphics[width=4.5in]{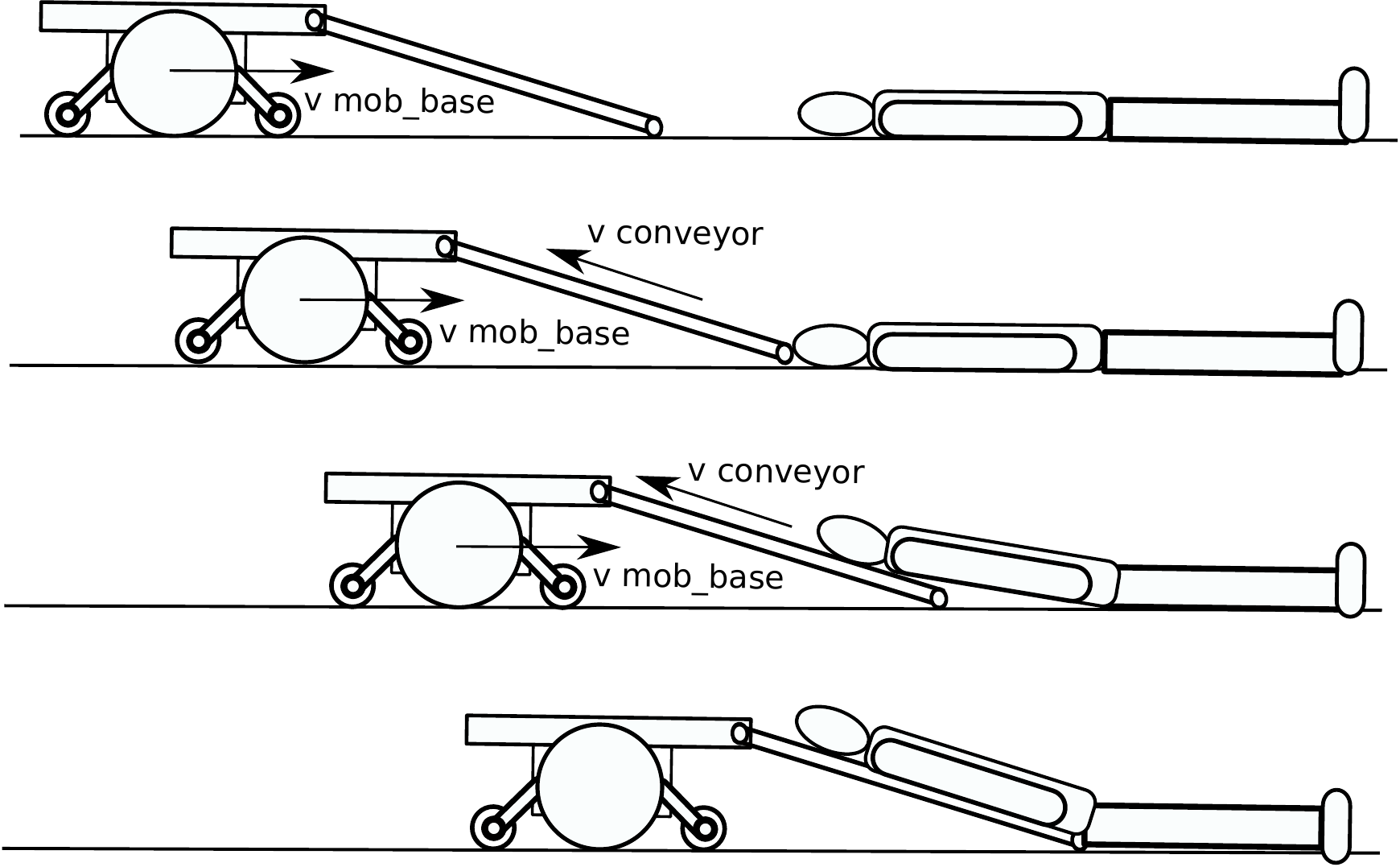}
%\vspace{-0.8em}
\caption{Illustration of casualty extraction via loco-manipulation technique. From the top to the bottom: 1) ResQbot gently approaching the victim, 2) the belt conveyor on the stretcher bed moving at the synchronized speed of the mobile
base, 3) the belt conveyor pulling the victim onto stretcher bed while the mobile base move to the opposite direction, 4) the victim fully onboard the stretcher bed.}
%\vspace{-1.0em}
\label{fig:method}
\vspace{-1.0em}
\end{figure}

Figure~\ref{fig:method} illustrates the proposed casualty-extraction operation using the loco-manipulation technique. By removing the need for high-complexity robotic manipulators or mechanisms, this technique greatly simplifies the underlying controls required for conducting complex casualty extraction in rescue missions. This simplicity is highly beneficial for intuitive teleoperation by human operators.

This casualty-extraction procedure involves four major phases:

\begin{enumerate}
\item \textbf{Relative pose adjustment}: the robot aligns its relative pose with respect to the victim in preparation for performing the loco-manipulation routine.
\item \textbf{Approaching}: the robot gently approaches the victim to safely make contact with the victim's head for initiating the loading process.
\item \textbf{Loading}: by using a balance between the locomotion of the base and the motion of the belt conveyor, the robot smoothly loads the victim onboard. Smooth operation at this stage is crucial in order to minimise traumatic injury caused by the operation. The victim is fully onboard when the upper body is fully loaded onto the stretcher bed. We consider the upper body to be from the head to the hip, thus protecting the critical parts of the body, including the head and spinal cord.
\item \textbf{Fastening}: once the victim is fully onboard, the strapping mechanism fastens the victim using a stretcher-strap mechanism in preparation for safe transportation. The conveyor surface serves as a stretcher bed for transporting the victim to a safe zone where paramedics can provide further medical care.
\end{enumerate}

\section{Robot Platform}

We have designed and built a novel mobile rescue robot platform called ResQbot as a proof of concept for the implementation of the proposed casualty-extraction procedure. Figure 1a shows the ResQbot platform that has been developed in this project. This platform is designed to be able to perform a casualty-extraction task based on the proposed loco-manipulation approach. This platform consists of two main modules: a differential-drive mobile base and a motorised stretcher-bed conveyor module. The platform is also equipped with a range of perception devices, including an RGB-D camera and a stereoscopic camera rig; these devices provide the perception required during the operation. We used an Xbox joystick controller for the operator interface to control the robot. We also proposed an immersive teleperception interface for this mobile robot using an HTC Vive virtual reality headset. This headset will provide teleperception for the operator by displaying a real-time visual image sent from the onboard stereo camera of the mobile robot. Figure~\ref{fig:prototype}b shows the operator teleoperates ResQbot using the proposed teleoperation and teleperception devices.

\subsection{Mobile Base Module}
The mobile base module used for the ResQbot platform is a differential drive module. This platform is a customised version of a commercially available powered wheelchair---Quickie Salsa-M---manufactured by Sunrise Medical\footnote{Quickie Salsa-M powered wheelchair by Sunrise Medical: www.sunrisemedical.co.uk}. This mobile base is chosen for its versatile design and stability owing to its original design purpose, which was to carry the disabled both indoors and outdoors. This platform has a compact turning circle while ensuring stability and safety through its all-wheel independent suspension and anti-pitch technology over rough or uneven terrain. Its mobile base is also capable of manoeuvring through narrow pathways and confined spaces, due to its compact design (only 600~mm in width).

\subsection{Motorised Stretcher Bed Module}

ResQbot is equipped with an active stretcher-bed module that enables active pulling up of the victim's body while the mobile platform is moving. This module is mounted at the back of the mobile base via hinges allowing it to fold (for compact navigation) or unfold (for loading and transporting the victim) on demand. This stretcher bed is composed of a belt-conveyor module that is capable of transporting a maximum payload of 100~kg at its maximum power. This belt conveyor is powered by a 240~VDC motor with 500~W maximum power. The motor is controlled through a driver module powered by a 240~VAC onboard power inverter, and the pulse-width modulation (PWM) control signal is used to control the motor's speed.

During loading of the victim's body, the active-pulling speed of the belt conveyor has to be synchronised with the locomotion speed of the mobile platform. Therefore, this module is equipped with a closed-loop speed control system to synchronise the conveyor speed and the mobile base locomotion speed. An incremental rotary encoder connected to the conveyor's pulley is used to provide speed measurements of the belt conveyor as feedback to the controller. Another incremental rotary encoder is connected to an omnidirectional wheel attached to the floor. This encoder provides measurements of the mobile base linear speed. The measured mobile base linear speed is used for the speed reference of the conveyor controller.

For a safe transportation process, the victim has to be safely placed onboard the stretcher bed. Thus, this stretcher-bed module is also equipped with a motorised stretcher strap to enable fastening of the victim on the bed as a safety measure. This stretcher-strap module is powered by a 24~VDC motor. The motor is controlled to fasten and unfasten the stretcher strap during the casualty-extraction procedure.

\section{Experimental Setups and Results}

\textbf{ Experimental Setting.} We have conducted a number of experimental trials to evaluate our proof-of-concept mobile-rescue-robot platform, ResQbot, in terms of task accomplishment, safety and teleperception comparison. In these experiments, we conducted a mock casualty-extraction procedure using ResQbot by teleoperation with three different teleperception modalities:

\begin{itemize}
\item \textbf{Direct mode (baseline)}: user controls the robot while being present at the scene;
\item \textbf{Conventional mode}: user receives visual feedback provided by a monocular camera through a display monitor;
\item \textbf{Immersive mode}: the user receives stereoscopic vision provided by an onboard stereoscopic camera module and through a virtual reality headset.
\end{itemize}

Ninety series of trials in total were conducted, with 30 series for each teleperception modality. We conducted this number of trials to capture any possible problems encountered during the trials. For the whole experiment, we used the same setup of the victim and its relative position and orientation with respect to the ResQbot. These various victim positions are inside the area of the ResQbot perception device's field of view.

To evaluate the safety of the proposed casualty-extraction procedure using the ResQbot platform, we conducted the trials using a sensorised dummy as the casualty. This dummy was equipped with an inertial measurement unit (IMU) sensor placed on its head. We used 3DM-GX4-25\footnote{LORD MicroStrain IMU: http://www.microstrain.com/inertial/3dm-gx4-25} IMU sensor, with resolution $<$ 0.1~mg and bias instability $\pm$ 0.04~mg. During the trials, we recorded the data from this sensor at 100~Hz sampling frequency.  

\begin{figure}[!t]
    \centering
        \subfloat[Relative pose adjustment] {\includegraphics[width=0.48\columnwidth]{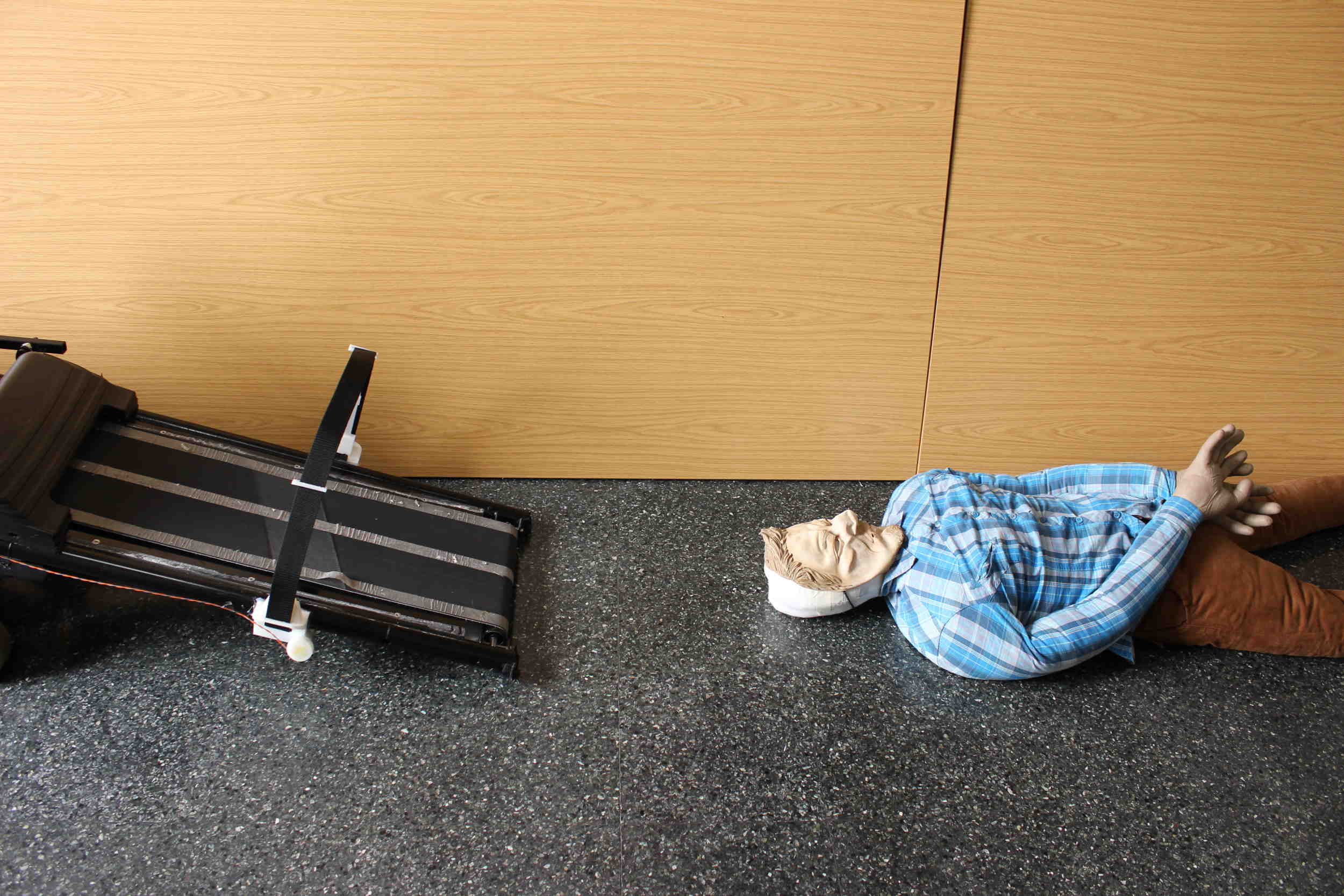}}
        \label{fig:resqbot-platform}
         \hspace{0.05cm}
        \subfloat[Approaching]{\includegraphics[width=0.48\columnwidth]{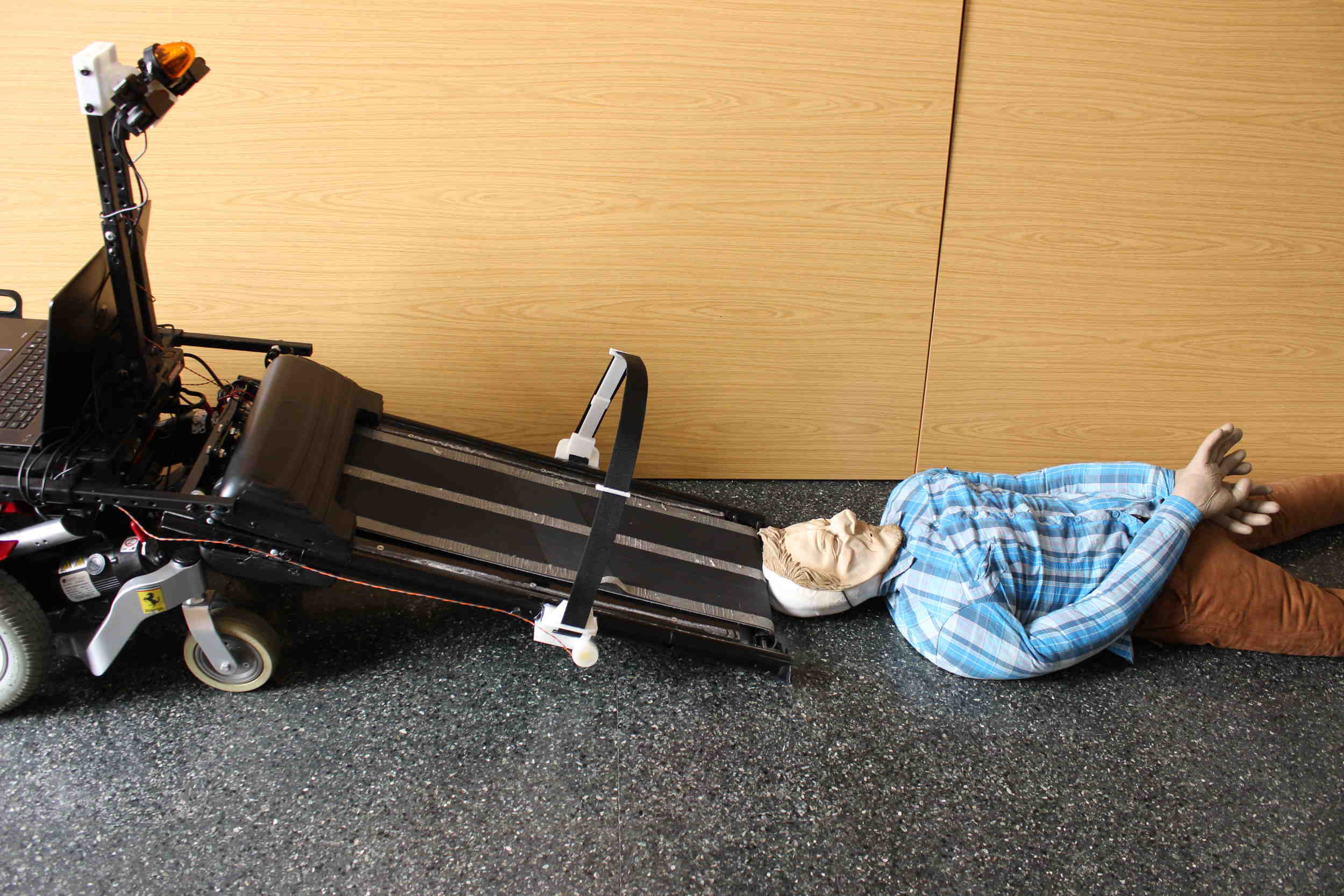}}
        \label{fig:resqbot-tele}
        \hspace{0.05cm}
        \subfloat[Loading] {\includegraphics[width=0.48\columnwidth]{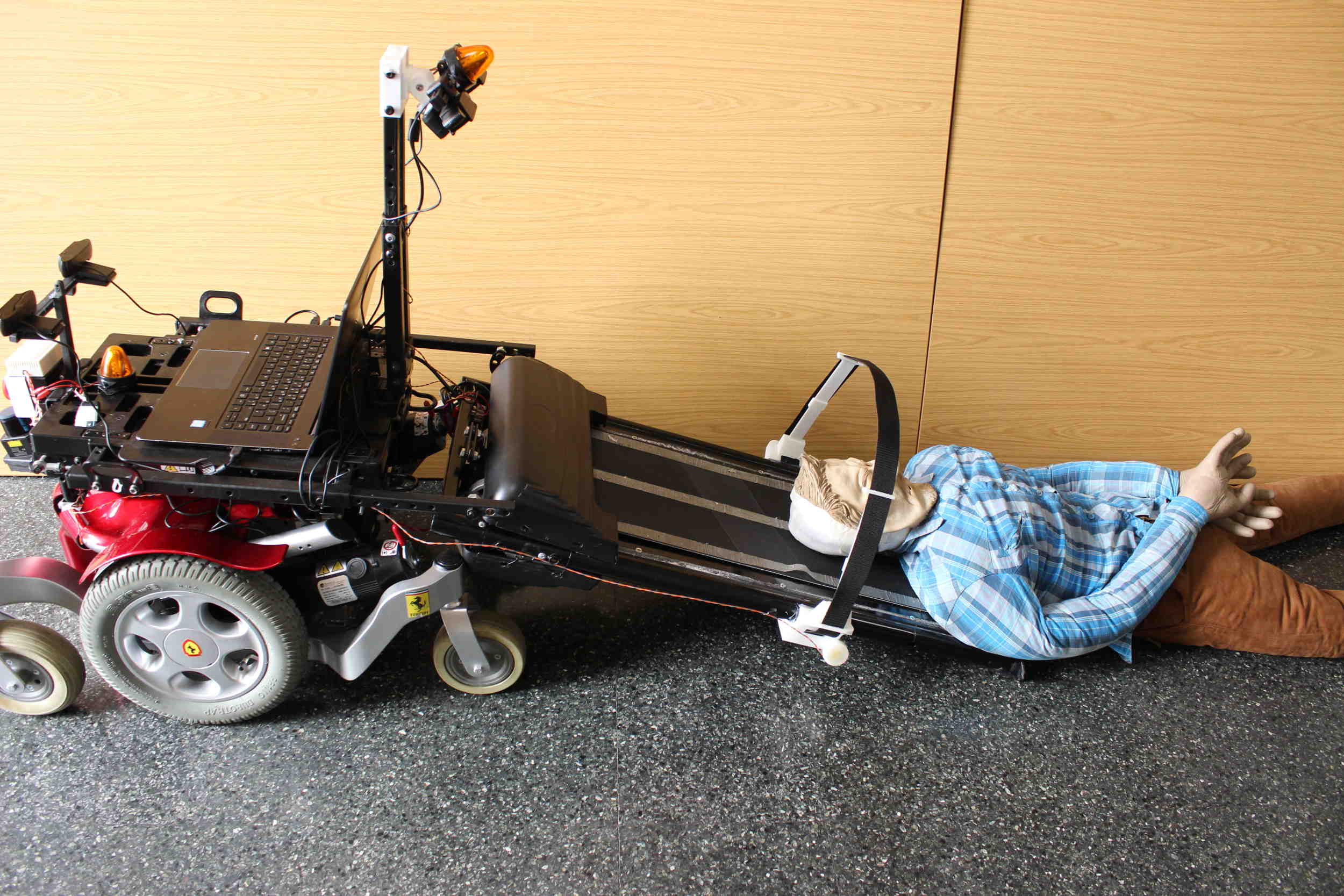}}
        \label{fig:resqbot-platform}
        \hspace{0.05cm}
        \subfloat[Fastening] {\includegraphics[width=0.48\columnwidth]{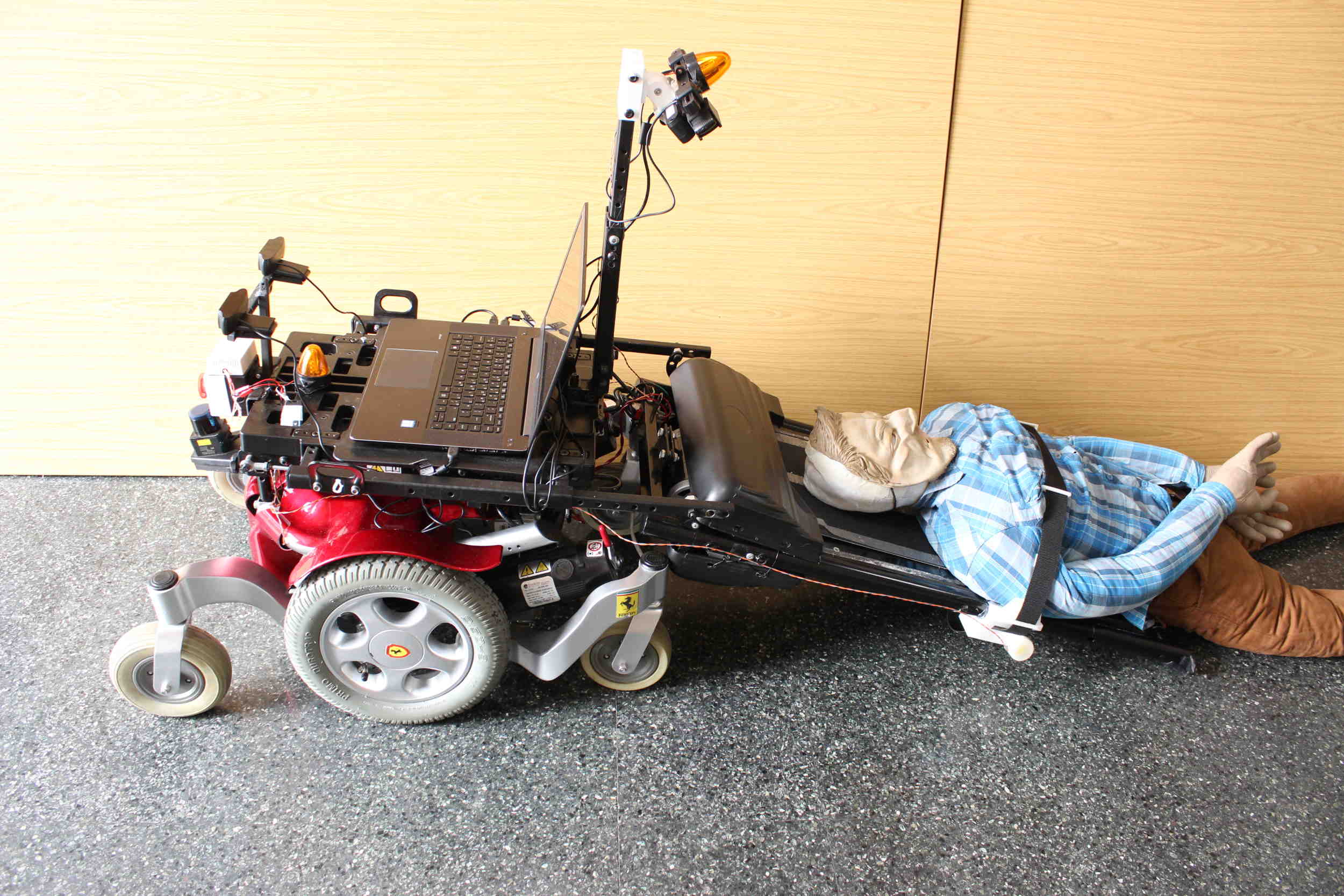}}
        \label{fig:resqbot-platform}         
    \caption{A sequence of images showing the progress of the casualty extraction task in the simulated rescue mission scenario in our experiments.}
    \label{fig:experiment}
\end{figure}

\textbf{Task Accomplishment}. In general, we achieved successful task accomplishment of the casualty-extraction procedure in every trial. Screenshot images in Figure~\ref{fig:experiment} demonstrate the procedure applied during the casualty-extraction operation performed by ResQbot in the experiments.

\textbf{Safety Evaluation}. In this experiment, we observed the impact applied to the dummy's head during the casualty-extraction procedure. This impact caused force and displacement of the dummy's head. The observation was focused on the part during the loading phase when the robot---i.e. the stretcher-bed conveyor---made first contact with the dummy's head. Two extreme cases were selected (i.e. roughest and smoothest operation, respectively) for this safety evaluation. These two cases represent the largest and the smallest maximum instant acceleration of each test within the whole experiment.

\begin{figure}[!t]
    \centering
        \subfloat[Dummy's head displacement] {\includegraphics[width=0.5\columnwidth]{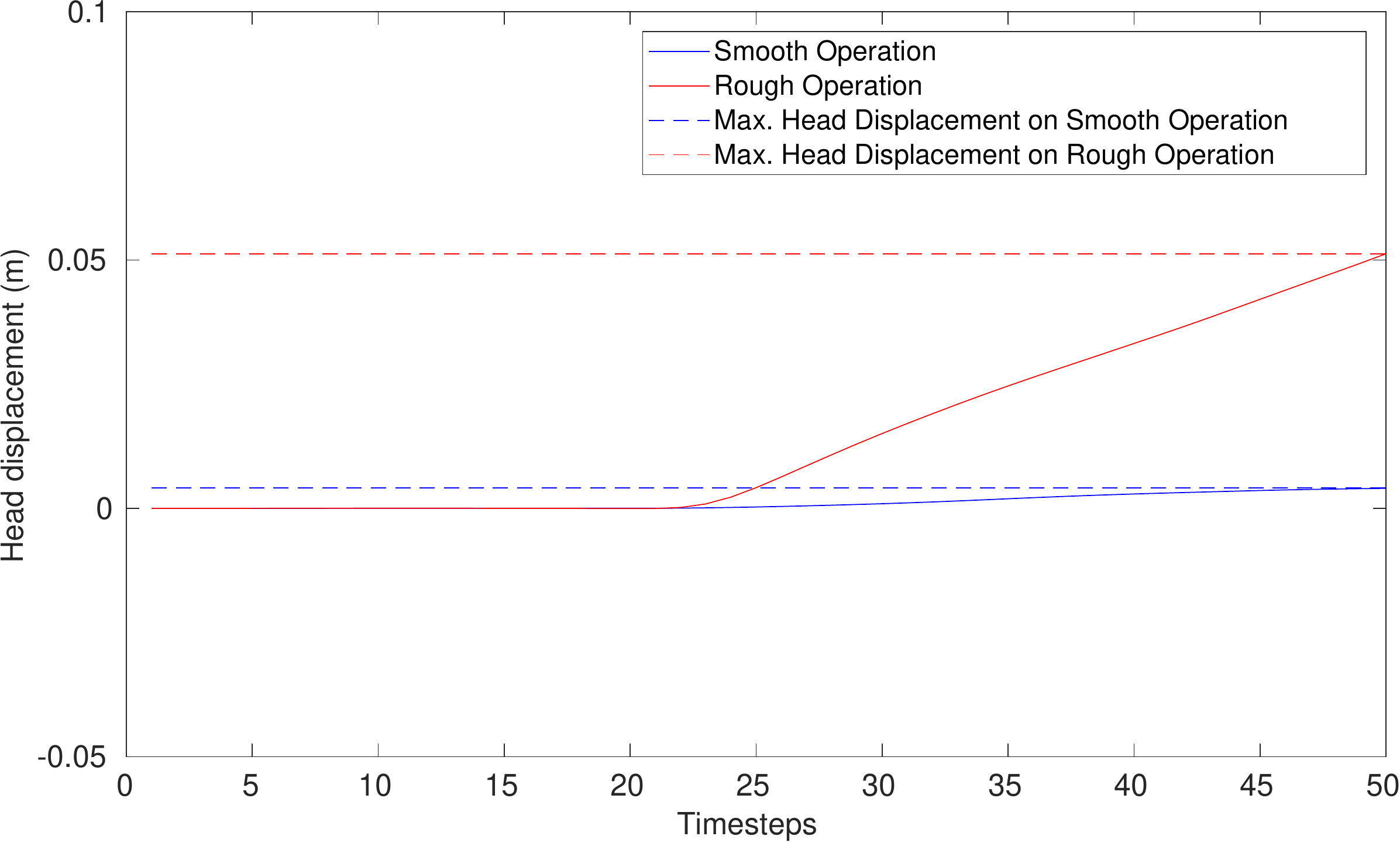}}
        \label{fig:resqbot-platform}
         \hspace{0.5cm}
        \subfloat[Dummy's head speed]{\includegraphics[width=0.46\columnwidth]{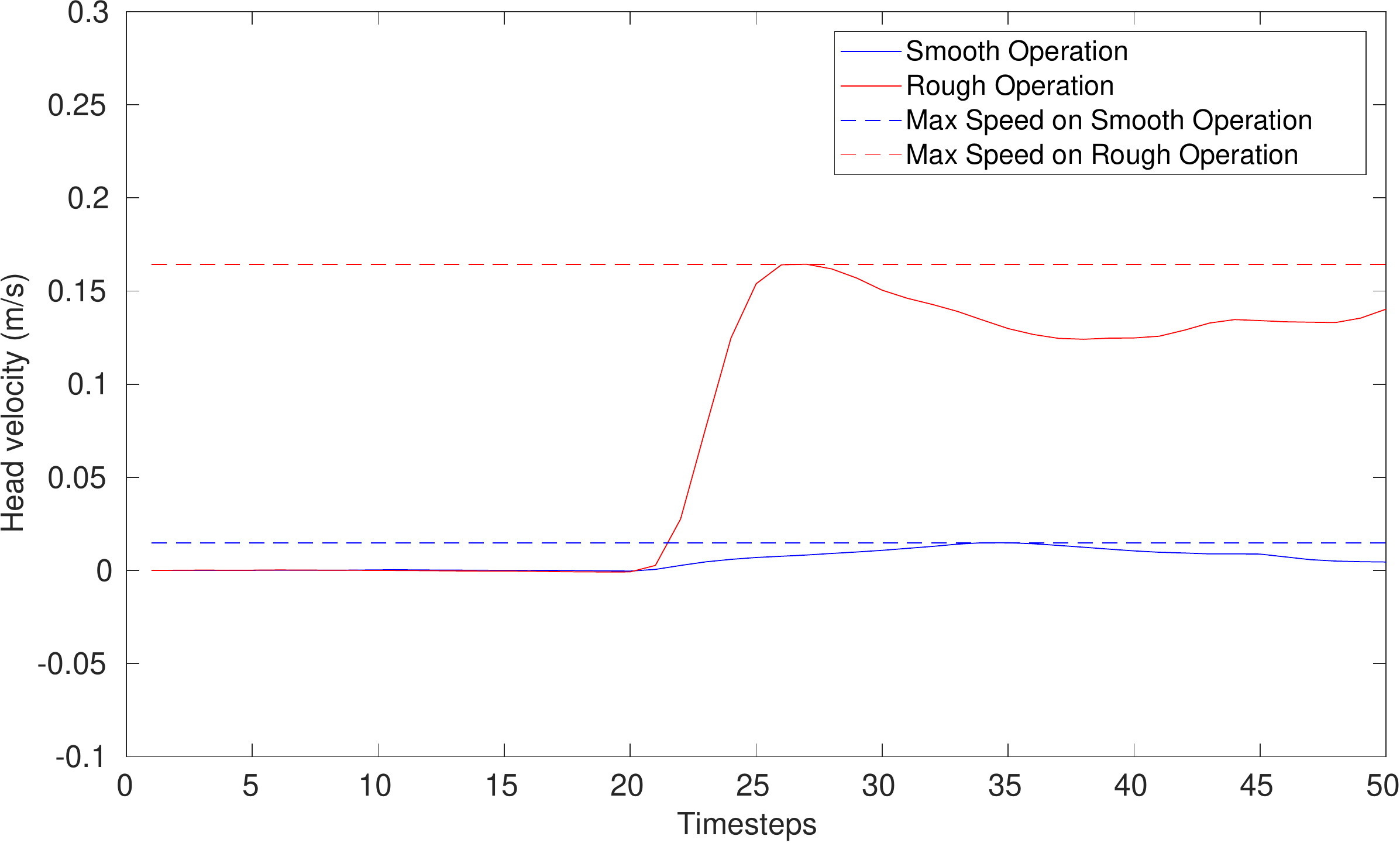}}
        \label{fig:resqbot-tele}
        \hspace{0.05cm}
        \subfloat[Instantaneous acceleration] {\includegraphics[width=0.48\columnwidth]{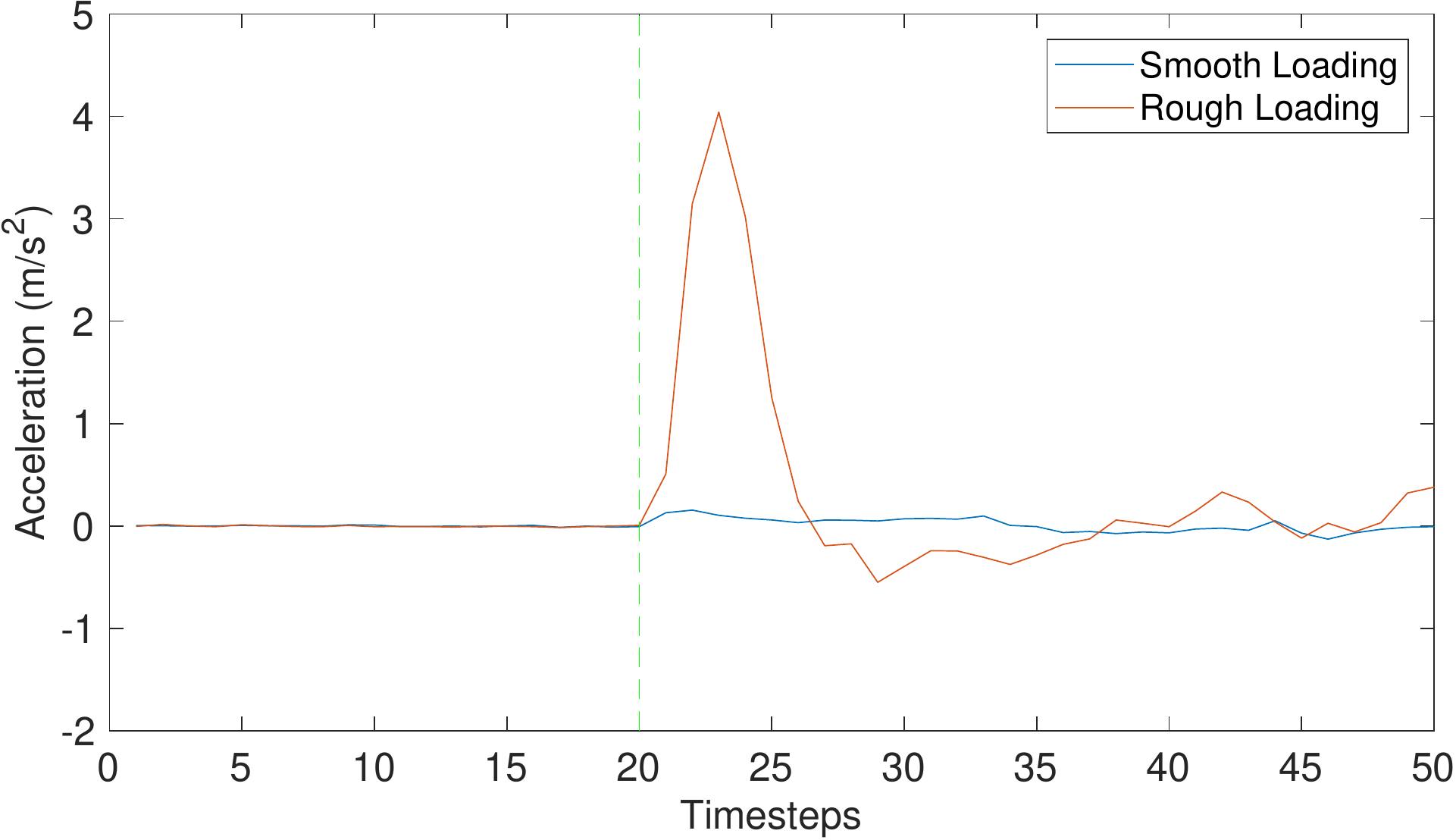}}
        \label{fig:resqbot-platform}
        \hspace{0.05cm}        
    \caption{The observation of two extreme cases of casualty extraction trial, (1) the smoothest trial (in blue) and (2) the roughest trial (in red). The dashed vertical line (in green) indicates the time-step at which the contact is initiated.}
    \label{fig:results}
    \vspace{-1.0em}
\end{figure}

Figure~\ref{fig:results} shows the dummy's head displacement, speed and acceleration during the loading process caused by the robot's first contact. Two extreme cases are presented; one is the smoothest trial (in blue), and the other one is the roughest trial (in red). The dashed vertical line (in green) indicates the time-step at which the contact was initiated during the loading process. 

The forces applied to the dummy's head can be estimated based on these measured accelerations from the IMU, and it can be calculated via: $$F_i=ma+F_s$$ 
in which, $F_i$ corresponds to the estimated instantaneous force applied to the dummy's head, and $F_s$ corresponds to the static friction force between the dummy's head and the ground~\cite{22:friction}. The estimated force was calculated based on the measured maximum instant acceleration (a) during the operation and the approximated mass of the dummy's head~(m). Table~\ref{tab1} summarises the observations of the two significant cases---i.e. smoothest and roughest trial---during the experiment. These cases correspond to the maximum instantaneous accelerations of the dummy's head caused by the robot during the casualty-extraction procedure.

\begin{figure}[t!]
\centering
\includegraphics[width=4.75in]{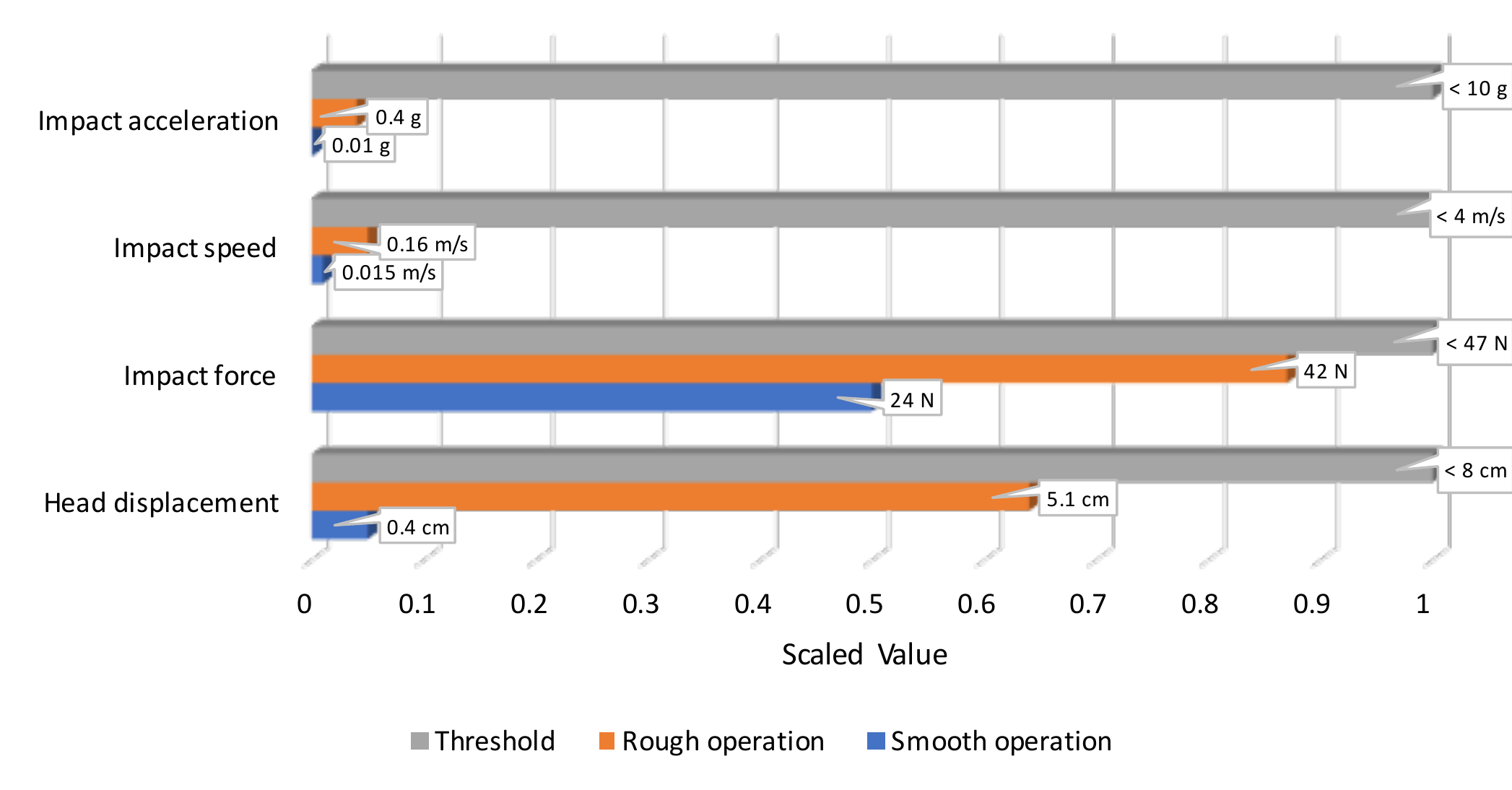}
\vspace{-0.8em}
\caption{The comparison between the trial results and the thresholds from the literature.}
\vspace{-1.0em}
\label{fig:summary}
\end{figure}

We compared the results presented in table~\ref{tab1} with several key safety thresholds---which were reported in the literature as possible causes of head or neck injuries to the casualty~\cite{23:Engsberg:2009spinal,24:eurailsafe:online}. Figure~\ref{fig:summary} illustrates the comparison between the trial results and the thresholds from the literature. It can be seen that all evaluated parameters in the experiments are relatively small and below the threshold. Even though not conclusive yet, these preliminary results show high safety promise for the proposed platform of the casualty-extraction procedure, and for further development. Thus it also motivates more elaborate safety evaluations for the practical deployment of the platform.

%\vspace{-1.0em}
\begin{table}
\caption{The summary of two significantly different trials---i.e. smooth and rough operation---performed during the experiment.}\label{tab1}
\vspace{1.0em}
\centering
\begin{tabular}{rrrrr}
\hline
 & \hspace{2.0em} &\textbf{Smooth Trial} & \hspace{2.0em} &\textbf{Rough Trial}\\
\hline
Max. instant acceleration~($m/s^2$) &  & $\approx 0.154$ & &$\approx 4.042$\\
Max. Velocity~($m/s$) \\ during initial contact &  &  $\approx 0.015$ & &$\approx 0.16$\\
Victim's head displacement~($m$) &  & $\approx 0.004$ & &$\approx 0.051$\\
Max. impact force~($N$) &  & $\approx 23.63$ & &$\approx 41.12$ \\
\hline
\end{tabular}
\end{table}
%\vspace{-1.0em}

%\vspace{-1.0em}

\textbf{Teleperception Comparison}. 
In this work, we proposed immersive teleperception via an HTC Vive headset to provide visual perception for the operator when operating the ResQbot platform. We evaluated this proposed teleperception modality by comparing it to the conventional teleperception setup (i.e. providing visual perception via single monitor) and the direct-observation scenario as a baseline (i.e. controlling the robot while being present at the scene).

\begin{figure}[t!]
\centering
\includegraphics[width=3.0in]{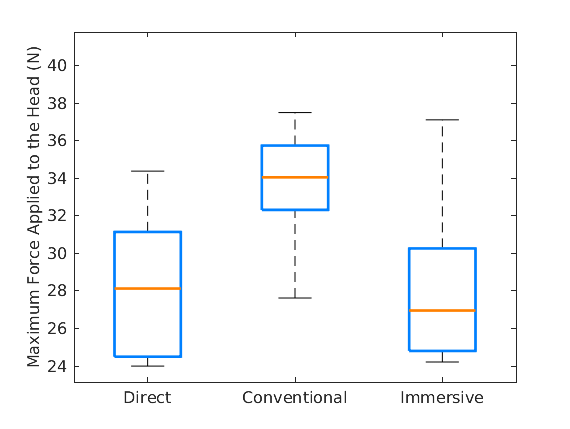}
%\vspace{-0.8em}
\caption{The distribution of the estimated maximal forces in the three different perception modes during the experiments.}
\vspace{-1.0em}
\label{fig:boxplot}
\end{figure}

We compared these three perception modes in terms of the smoothness of the casualty-extraction procedure, which is represented by the maximum force applied to the dummy's head during the experiment in both cases. Figure~\ref{fig:boxplot} shows a box plot of the distribution of the estimated maximal forces in the three different perception modes during the experiments. According to Figure~\ref{fig:boxplot}, in terms of the smoothness of the procedure, we observed that the VR mode (i.e. immersive) results in a higher population of smoother trials than the conventional mode. In fact, the VR mode achieves smooth operations with low maximal estimated forces under 28 N for approximately 50$\%$ of its trials, similar to the direct-observation mode, in which the operator has direct access to observe the scene during the operation.

\section{Conclusion}

In this paper, we presented a proposed mobile rescue robot system that is capable of safely loading and transporting a casualty. We proposed an immersive stereoscopic teleperception modality via an HTC Vive headset to provide the teleoperator with more realistic and intuitive perception information during the operation. As a proof of concept of the proposed system, we designed and built a novel mobile rescue robot called ResQbot, which is controlled via teleoperation. We evaluated the proposed platform in terms of task accomplishment, safety and teleperception comparison for completing a casualty-extraction procedure. Based on the results of our experiments, the proposed platform is capable of performing a safe casualty-extraction procedure and offers great promise for further development. Moreover, the teleperception comparisons highlight that the proposed immersive teleperception can improve the performance of the teleoperator controlling the mobile robot's performance during a casualty-extraction procedure.

% \begin{figure}[t!]
% \centering
% \includegraphics[width=3.9in]{method-crop.pdf}
% %\vspace{-0.8em}
% \caption{The proposed methodology for casualty detection via ground-projected point-cloud image.}
% %\vspace{-1.0em}
% \label{fig:method}
% \vspace{-1.0em}
% \end{figure}

% % 
% \begin{figure}[!t]
%     \centering
%         \subfloat[RGB image]{\fbox{\includegraphics[width=0.31\columnwidth]{rgb-image.png}}}
%         \label{fig:resqbot-platform}
% %         \hspace{0.1cm}
%         \subfloat[Point-cloud data and detected plane]{\fbox{\includegraphics[width=0.31\columnwidth]{point-cloud.png}}}
%         \label{fig:resqbot-tele}
% %         \subfloat[]{\fbox{\includegraphics[width=0.21\columnwidth]{point-cloud.png}}}
% %         \label{fig:resqbot-tele}
%         \subfloat[GPPC image and detected human body]{\fbox{\includegraphics[width=0.26\columnwidth]{detected.jpg}}}
%         \label{fig:resqbot-tele}
%         %\quad 
%         \vspace{-1.0em}
%     \caption{Preliminary experimental results from casualty detection using the proposed method.}
%     \label{fig:result}
%     \vspace{-1.6em}
% \end{figure} 

\vspace{-0.8em}
\section*{Acknowledment}
\vspace{-0.8em}
Roni Permana Saputra would like to thank Indonesia Endowment Fund for Education - LPDP, for the financial support of the PhD program.  
The authors would also like to show our gratitude to Arash Tavakoli and Nemanja Rakicevic for helpful discussions and inputs for the present work. 
\vspace{-0.8em}
%%%%%%%%%%%%%%%%%%%%%%%%%%%%%%%%%%%%%%%%%%%%%%%%%%%%%%%%%%%%%%%%%%%%%%%%%%%%%%%%%%%%


\end{document}